\documentclass[acmsmall]{acmart}

\usepackage{graphicx}
\usepackage{subfigure}
\usepackage{booktabs} 
\usepackage[T1]{fontenc}
\usepackage[ruled]{algorithm2e} 

\usepackage{flushend}

\AtBeginDocument{%
  \providecommand\BibTeX{{%
    \normalfont B\kern-0.5em{\scshape i\kern-0.25em b}\kern-0.8em\TeX}}}

\acmConference[]{}{}{}

\begin{document}

\title{A Reality check of the benefits of LLM in business}


\author{Ming Cheung}
\affiliation{%
  \institution{Beta Labs, The Lane Crawford Joyce Group}
  \country{China}
}
\email{mingcheung@lcjgroup.com}




\begin{abstract}
Large language models (LLMs) have achieved remarkable performance in language understanding and generation tasks by leveraging vast amounts of online texts. 
Unlike conventional models, LLMs can adapt to new domains through prompt engineering without the need for retraining, making them suitable for various business functions, such as strategic planning, project implementation, and data-driven decision-making. 
However, their limitations in terms of bias, contextual understanding, and sensitivity to prompts raise concerns about their readiness for real-world applications. 
This paper thoroughly examines the usefulness and readiness of LLMs for business processes. 
The limitations and capacities of LLMs are evaluated through experiments conducted on four accessible LLMs using real-world data. 
The findings have significant implications for organizations seeking to leverage generative AI and provide valuable insights into future research directions. 
To the best of our knowledge, this represents the first quantified study of LLMs applied to core business operations and challenges.
\end{abstract}




\maketitle

\section{Introduction}
\noindent
Recent advancements in artificial intelligence (AI), particularly natural language processing (NLP) techniques, have yielded powerful new tools with significant implications for business applications. 
Specifically, modern AI systems equipped with NLP capabilities exhibit the potential to understand, analyze, and generate human language. 
For industry, these developments open new opportunities to gain insights from vast amounts of unstructured data sources, such as comments from social media. 
NLP models demonstrate the ability to extract key elements from documents and other text data, ranging from assessing sentiment in customer feedback to summarizing lengthy reports. 
However, effectively utilizing NLP techniques requires expertise in NLP, including data preparation and model training, which limits its accessibility to experts.
\\
\indent
One area of particular promise is the development of large language models (LLMs). 
These models, such as ChatGPT from OpenAI, have revolutionized the applications of AI. 
LLMs are known for their ability to generate coherent and contextually relevant responses.
Unlike conventional NLP models, LLMs can be customized for specific applications using prompt engineering \cite{ouyang2022training}, eliminating the need for retraining or additional data. 
LLM models contain hundreds of billions of parameters, far surpassing the size of conventional NLP models. 
This allows them to be trained on trillions of words of text from the Internet, enabling them to capture broad linguistic patterns. 
Consequently, LLMs offer possibilities that extend beyond traditional NLP tasks.
While LLMs show promise for various tasks, it remains unclear how effectively they can assist with other business functions. 
The limitations of bias, contextual understanding, and sensitivity to prompts in LLMs raise questions about their readiness for real-world applications and the specific aspects in which LLMs can assist human work.
\\
\indent
This paper serves as a reality check to assess the potential of LLMs in assisting with business processes. 
It discusses the development and common types of LLMs, followed by an exploration of their business value. 
The limitations of LLMs are also discussed, and experiments are conducted to verify their effects using real data from four accessible LLMs. 
Prompt examples in this paper are generated by ChatGPT-3.5 through POE \footnote{available: https://poe.com/ChatGPT} if not specified.
The main contributions of this paper are as follows:
\begin{itemize}
\item Discussion of common LLMs and their use cases for business projects.
\item Evaluation of the limitations of LLMs, including bias, contextual understanding, and sensitivity to prompts.
\item Conducting extensive experiments to verify the usefulness and capacities of LLMs for business using real data from four accessible LLMs.
\end{itemize}

The paper is organized as follows: Section \ref{sec:related_works} discusses related works, and Section \ref{sec:common_llms} introduces common LLMs. Section \ref{sec:LLM_for_Business} explores the application of LLMs in business processes. Sections \ref{sec:bias}, \ref{sec:no_context}, and \ref{sec:sensitivity} discuss the limitations of LLMs with experiments using real data. Finally, Section \ref{sec:conclusion} concludes the paper.

\section{Related Works}
\label{sec:related_works}
\begin{figure*}
\centering 
\includegraphics[width=5.5in]{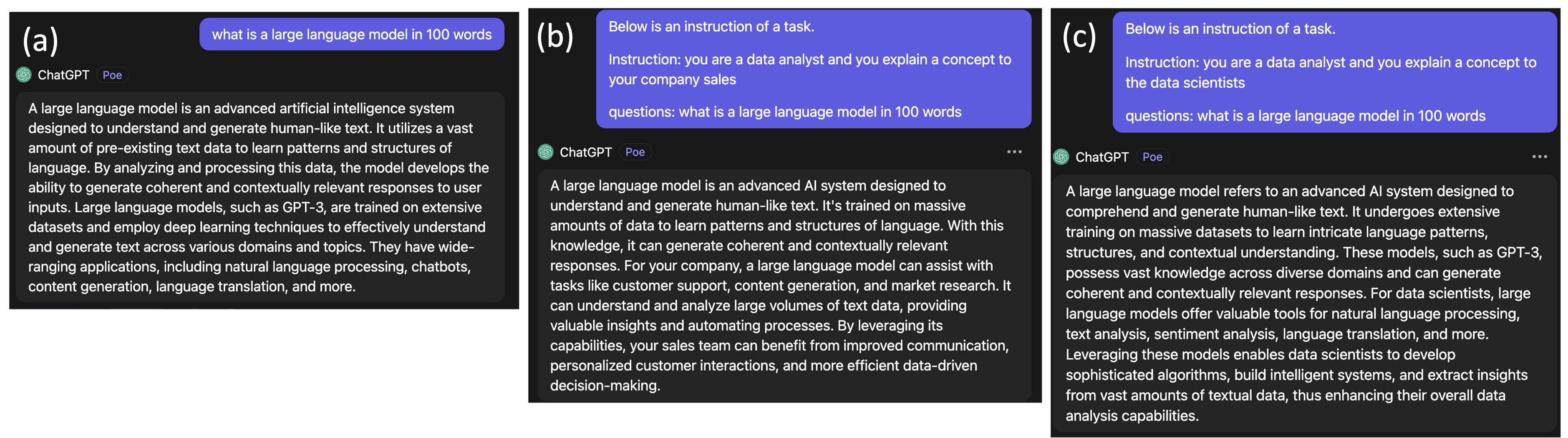} 
\caption{Examples of a question on LLM with different prompts: (a) the question only; (b) write for sales, (c) write for data scientists.}
\label{fig:prompt_enginnering}
\end{figure*}

\noindent
This section introduces related works on LLMs, covering their developments and prompt engineering.
The development of LLMs has undergone significant advancements recently.
Conventional NLP models, such as recurrent neural networks (RNNs) \cite{salehinejad2017recent} and long short-term memory (LSTM) models \cite{graves2012long}, paved the way for understanding sequential data, especially texts. 
However, with the introduction of the transformer architecture \cite{gillioz2020overview}, models became capable of capturing long-range dependencies in texts.
The transformer architecture employs self-attention mechanisms \cite{vaswani2017attention} that enable it to focus on different parts of the input text and generate coherent and contextually relevant responses.
As a result of the transformer's development, many LLMs have been created, such as GPT-3 \cite{dale2021gpt}, Google Bard \cite{manyika2023overview}, and NeevaAI \cite{liu2023evaluating}. 
These models, trained with human feedback \cite{ouyang2022training}, consist of at least billions of parameters and can understand and respond to prompts in a conversational manner \cite{brown2020language}. 
Consequently, they have found applications in chatbots, virtual assistants, and natural language understanding tasks.
Unlike conventional models that require labeled data and retraining for new tasks, LLMs can generate human-like responses \cite{kaddour2023challenges} for new tasks without the need for retraining, through prompt engineering. 
LLMs exhibit impressive capabilities in a range of NLP tasks, including language translation \cite{reynolds2021prompt}, sentiment analysis, and named entity recognition \cite{kaddour2023challenges, xie2023translating}.
Fig. \ref{fig:prompt_enginnering} (a) shows an example of asking an LLM the question: "What is a Large Language Model in 100 words?" The LLM provides a clear explanation suitable for the general public. 
Figures \ref{fig:prompt_enginnering} (b) and (c) demonstrate the same question with different prompts, resulting in different responses based on the instructions in the prompts. 
It is interesting to investigate the potential of LLMs when provided with suitable prompts.
\\
\\
\indent
Prompt engineering plays a crucial role in leveraging the capabilities of the LLM by providing specific instructions, context, or background information about a task, along with the associated questions or expected outputs \cite{zhou2023context}. 
By utilizing prompt engineering, LLM can generate texts by drawing from existing websites or documents, rather than generating content entirely from scratch. 
As shown in Fig. \ref{fig:prompt_enginnering}, the presence or absence of a prompt significantly impacts the generated text's style and suitability for different audiences. 
For example, the prompt designed for sales purposes employs simpler language and has broader applications, making it more accessible to laypersons in the sales field. 
In contrast, Fig. \ref{fig:prompt_enginnering} (c) showcases a prompt tailored for data scientists, incorporating more specialized terms like sentiment analysis. 
This demonstrates the power of prompt engineering in shaping LLM's output. 
Notably, prompt engineering enhances LLM's contextual faithfulness and allows for behavioral constraints \cite{reynolds2021prompt}, enabling a wide range of tasks, including data labeling \cite{bonifacio2022inpars, ding2022gpt}, data generation \cite{dai2023chataug}, recommendation \cite{gao2023chat, wang2022towards} and code generation \cite{biswas2023role, chen2023teaching}. 
LLM has also proven its versatility across various domains, such as generating textbook questions \cite{jalil2023chatgpt}, fact-checking \cite{lee2020language}, and facilitating logical thinking \cite{zhou2020evaluating}. 
While self-feedback techniques \cite{madaan2023self} can further improve the results, expanding LLM's capabilities by incorporating different modules, such as images \cite{ramesh2021zero} and audio \cite{sushko2021one}, shows promise. 
However, it is worth noting that while LLMs offer benefits in developing outlines and conducting literature reviews \cite{huang2023role}, the generated texts may lack support \cite{alafnan2023chatgpt}. 
Additionally, the customization and specialization of LLMs are limited, particularly for extensive fine-tuning on larger models, impeding their potential applications in areas like recommendation systems \cite{kanagavelu2020two}. 
Moreover, LLMs are vulnerable to carefully crafted adversarial examples, leading to incorrect or nonsensical outputs \cite{liu2023robustness}, although ongoing research aims to enhance LLM robustness against such attacks. 
Despite these challenges, the specific utility of LLMs in business contexts remains an open question, making this paper a reality check for assessing the potential of LLMs in the business domain.
\\
\begin{figure*}
\centering 
\includegraphics[width=5.5in]{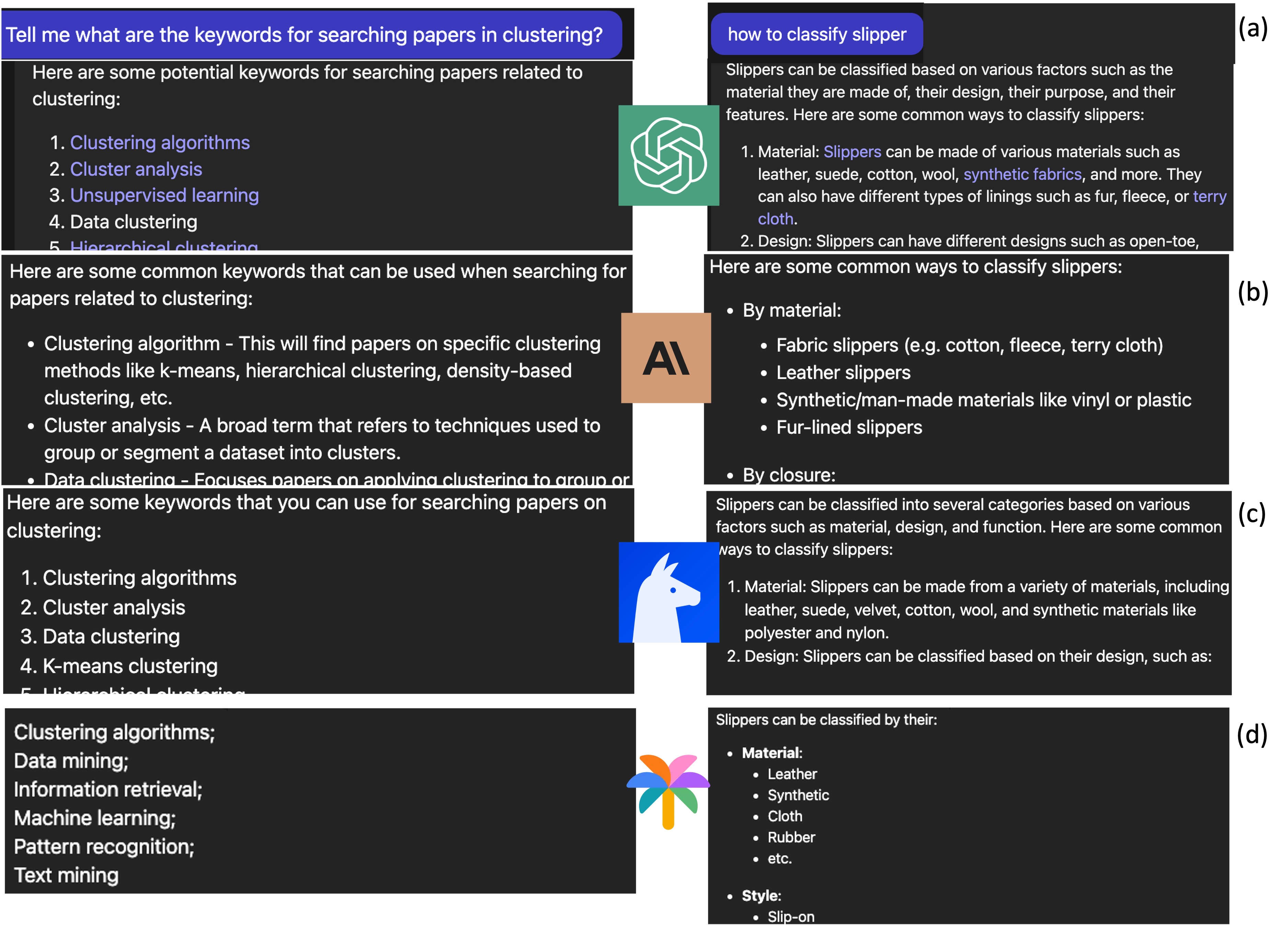} 
\caption{Example of questions on different LLM: (a) ChatGPT; (b) Claude; (c) Llama; (d) PaLM}
\label{fig:diff_llm}
\end{figure*}
\\
\section{Common LLMs}
\label{sec:common_llms}
\noindent
There are many different LLMs available, and due to the model structures and training data, they could generate different responses for the same prompt and question due to the different parameters, structures and training data.
Fig. \ref{fig:diff_llm} shows the responses of 2 questions on different LLMs.
The first question is about the keywords to be searched for papers in clustering while the second one about how slippers can be classified.
The first one requires general concepts and the second one requires some domain knowledge.
It is observed that all LLMs are able to generate meaningful answers, but some of them, such as Claude, provides more details, while PaLM, only provides the keywords.
This section introduces some more accessible LLMs that are released by big companies, such as, OpenAI, Facebook and Google.
They are good representatives of LLM in the market.

\subsection{ChatGPT 3.5}
\noindent
ChatGPT 3.5\footnote{https://openai.com/} is a conversational AI model developed by OpenAI. 
It is based on the GPT (Generative Pre-trained Transformer)\cite{radford2018improving} architecture and is designed to generate human-like responses in a conversational context. ChatGPT can be used for a wide range of tasks, including chatbots, virtual assistants, and dialogue systems. It is known for its ability to generate coherent and contextually relevant responses.
It is trained on internet-sourced data, that is, 570 gigabytes of text, including text from Wikipedia and Twitter.
The model has 175 billion parameters.

\subsection{Claude Instant}
Claude Instant \footnote{https://www.anthropic.com/} aims to be a faster and cheaper version of Claude, which is developed by Anthropic. 
They aim to design a safer LLMs.
It can handle a range of tasks including casual dialogue, text analysis, summarization, and document comprehension. 
Compared to the previously available Claude, Claude-instant is faster and significantly better at non-English languages. 
However, there is no available information on the training data and the number of parameters.

\subsection{Llama-2-70b}
Llama\footnote{https://huggingface.co/meta-llama/Llama-2-70b-chat-hf} is a LLM developed by Meta. 
It is one of the models available in the Llama series and is trained using the GPT-3.5 architecture. 
Llama-2-70b is designed to generate high-quality responses and can be fine-tuned for specific tasks. It can be used for a wide range of applications, including chatbots, content generation, and natural language understanding tasks\cite{touvron2023llama}.
There are 70 billion in the model, and it is trained on 2 trillion tokens of data from publicly available sources \cite{huggingFace}.

\subsection{PaLM 2}
PaLM is a LLM developed by Google AI \footnote{https://ai.google/discover/palm2}.
PaLM 2 chatbison model is the next generation of the PaLM model and is designed to excel at advanced reasoning tasks, including code, math, classification, question answering, translation and multilingual proficiency, and natural language generation better than its predecessor. 
PaLM 2 was pre-trained on parallel multilingual text and on a much larger corpus of different languages than its predecessor, PaLM. 
All versions of PaLM 2 are evaluated rigorously for potential harms and biases, capabilities and downstream uses in research and in-product applications\cite{anil2023palm}.
It is trained on web pages, research papers, books, code, and mathematical and conversational data \cite{palm2}.
However, the exact number of parameters used to train PaLM 2 is not publicly disclosed by Google.

\begin{figure*}
\centering 
\includegraphics[width=5.5in]{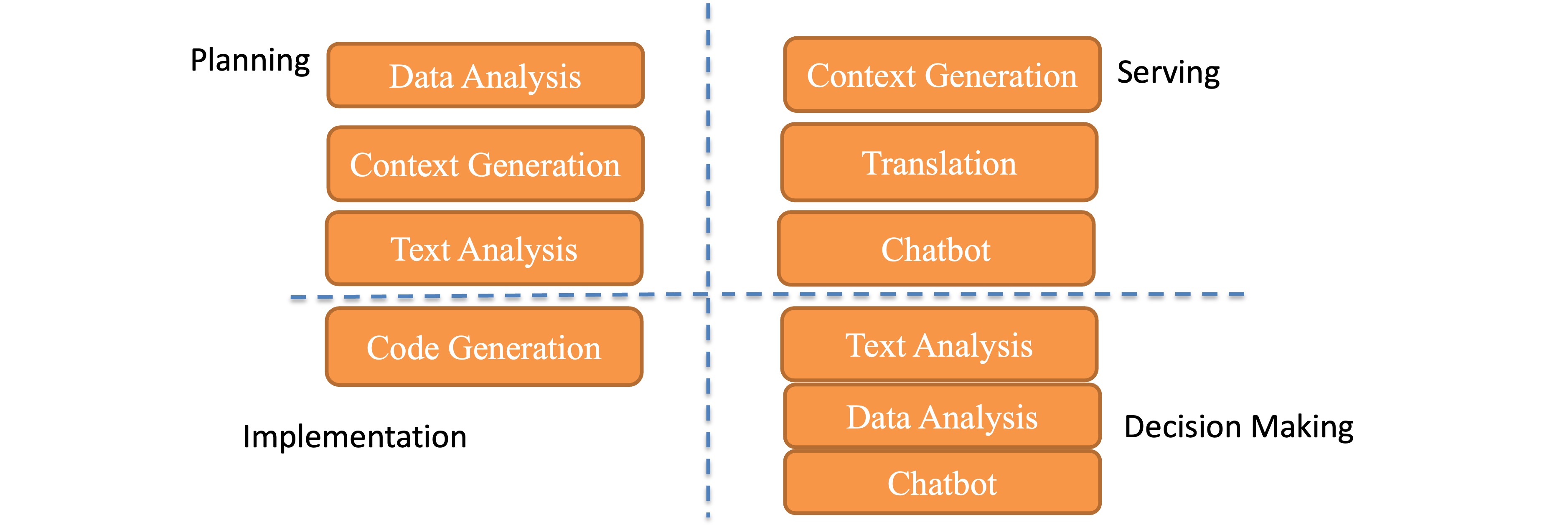} 
\caption{Phrases of using LLM for business project planning, from planning, implementation, to decision making and serving}
\label{fig:flow_business}
\end{figure*}

\section{LLM for Business}
\label{sec:LLM_for_Business}
The section discusses the diverse applications and capacity of pre-trained LLM that can revolutionize planning, implementation, serving and decision making.
Fig. \ref{fig:flow_business} shows different phrases of using LLM, including planning, implementation, serving and decision making.

\begin{figure*}
\centering 
\includegraphics[width=5.5in]{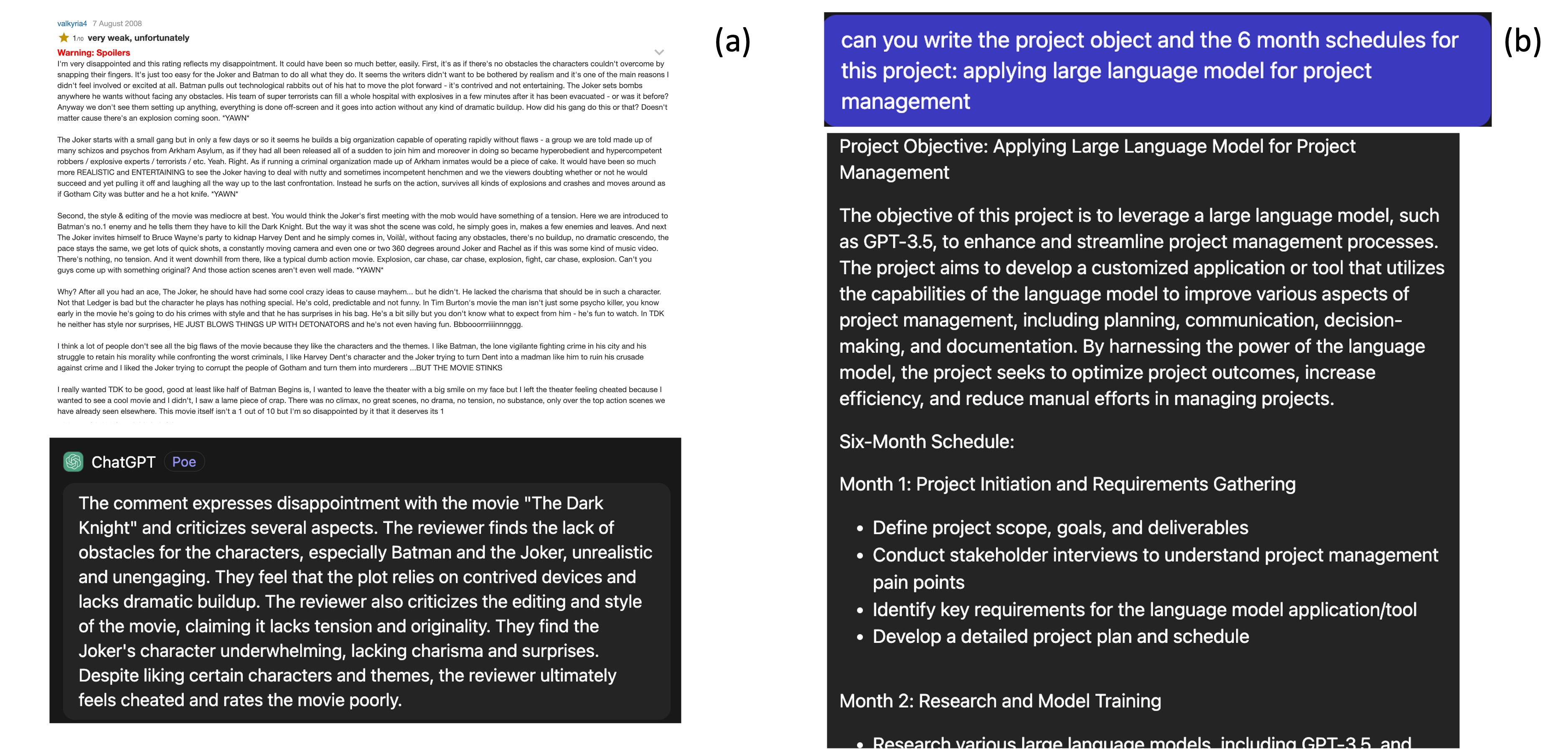} 
\caption{Examples of using LLMs for: (a) Text Analysis; (b) Content Generation.}
\label{fig:llm_text_content}
\end{figure*}

\subsection{Text Analysis}
\noindent
Text analysis plays a vital role in various business applications, particularly in planning and decision-making processes. 
Leveraging LLMs, such as ChatGPT \cite{huang2023chatgpt}, enables comprehensive analysis and understanding of the sentiment expressed in text, providing valuable insights into customer opinions and overall sentiment without the need for LLM retraining. 
Furthermore, LLMs have the ability to summarize and condense lengthy texts, making them easier to comprehend and significantly reducing the time required to process large amounts of information. 
This capability proves beneficial in project planning, as it allows for the understanding and summarization of existing projects, while also suggesting potential directions for new endeavors. 
When making business decisions concerning ongoing projects, analyzing customer comments empowers stakeholders to make appropriate improvements or even consider halting the project if necessary. 
Fig. \ref{fig:llm_text_content} (a) illustrates text analysis, showcasing an example where user comments about a movie\footnote{Available at: https://m.imdb.com/title/tt0468569/reviews} are analyzed. 
Even for lengthy inputs, the comment can be summarized into 50 words based on the provided prompt, enabling users to understand that it is a negative comment and the underlying reason behind it. 
By conducting such analyses on all comments, stakeholders can gain a comprehensive overview, expediting their understanding of a large volume of text.

\subsection{Content Generation}
\noindent
As depicted in Fig. \ref{fig:flow_business}, content generation plays a crucial role in planning and serving. \
Particularly, LLMs demonstrate remarkable abilities in generating tailored project plans by incorporating instructions, key points, and constraints into the prompt. 
This includes tasks such as formulating project objectives and creating schedules. 
Additionally, LLMs prove to be invaluable in serving scenarios by producing captivating social media posts and persuasive product descriptions, thereby attracting a larger customer base. 
In Fig. \ref{fig:llm_text_content} (b), an example showcases the generation of project objectives and schedules based solely on the project name. 
Notably, the generated information encompasses the desired objectives as well as a comprehensive monthly plan. 
By streamlining the project planning process, LLMs can significantly reduce the time and effort invested by professionals, enabling them to focus on their core areas of expertise \cite{lecler2023revolutionizing}.

\begin{figure*}
\centering 
\includegraphics[width=5.5in]{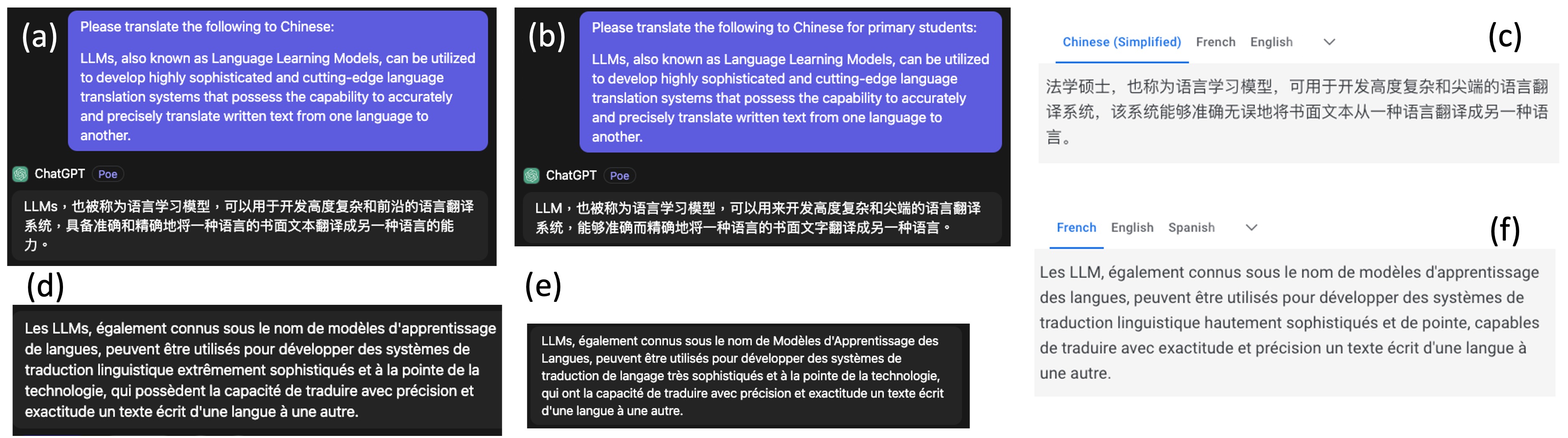} 
\caption{Examples of translations of the same texts: (a) to Chinese; (b) to Chinese with Prompt; (c) to Chinese using Google Translate; (d) to French; (e) to French with Prompt; (f) to French using Google Translate.}
\label{fig:llm_translate}
\end{figure*}

\subsection{Translation}
\noindent
Translation facilitates the conversion of texts into various customer languages, such as translating websites and product descriptions from English to Chinese. 
It offers significant benefits in terms of expanding the customer base, as illustrated in Fig. \ref{fig:flow_business}. 
By harnessing the capabilities of LLMs, organizations can achieve highly accurate and precise translations that are customizable, reliable, and of the utmost quality \cite{ghazvininejad2023dictionary}. 
LLMs enable businesses to overcome language barriers across diverse customer groups and cultures. 
In fact, LLMs have been shown to surpass Google Translate in performance and support a wide range of languages \cite{anil2023palm}. 
Fig. \ref{fig:llm_translate} provides examples of translations from English to Chinese and French. 
In particular, Fig. \ref{fig:llm_translate}(a) and (b) showcase translations into Chinese using different prompts, with and without specific requirements for primary students. 
Similarly, Fig. \ref{fig:llm_translate}(d) and (e) demonstrate translations into French using a similar prompt. 
Notably, the translations with the requirement to cater to a primary student are simpler and easier to comprehend. 
On the contrary, while Google Translate\footnote{Available: translate.google.com} can translate texts, it lacks customization capabilities and may introduce errors, such as mistranslating "LLM" as another subject. 
However, LLMs provide users with the flexibility to input specific terms in the prompt, allowing for accurate translations even for less common terms. 
This example illustrates how LLMs, through translation prompt engineering, empower businesses to control the content and style of translations to align with their specific use cases.

\begin{figure*}
\centering 
\includegraphics[width=5.5in]{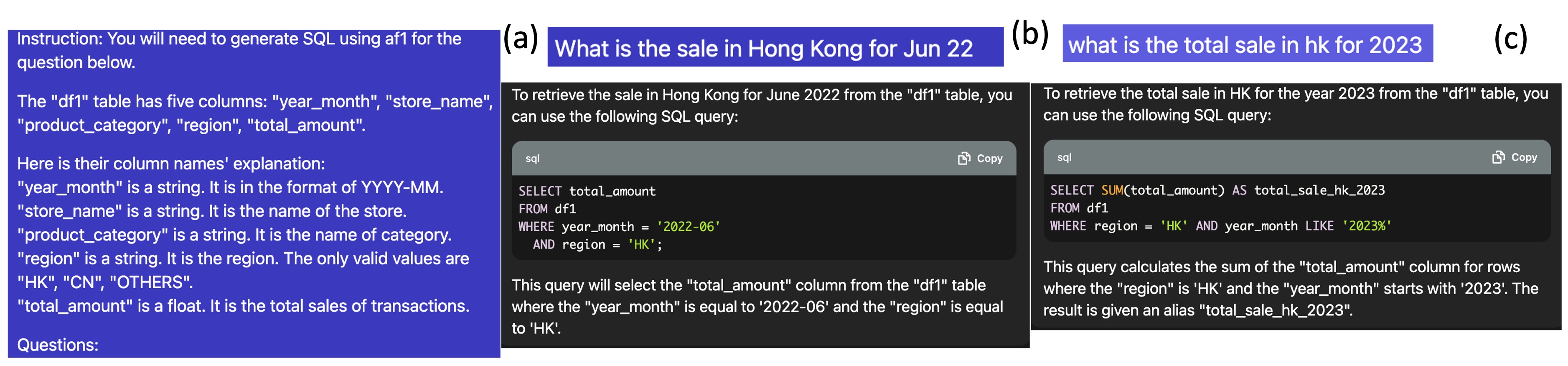} 
\caption{An example of code generation using LLM.}
\label{fig:Code_example}
\end{figure*}

\subsection{Code Generation}
Code generation plays a crucial role in project implementation, offering significant benefits. 
Recent advancements in LLMs have showcased their immense potential in automating diverse coding tasks, including generating code snippets or complete programs\cite{biswas2023role}. 
Among these programming tasks, SQL holds particular significance in the realm of business operations as it enables the extraction and comprehension of valuable insights from databases, such as transactions, customers, and retail store data. 
Traditionally, business analysts manually create SQL code based on specific business questions and requirements, which can be time-consuming for engineers and analysts to understand the question, as well as write the corresponding code to extract the right data. 
In contrast, LLMs can generate the required code within seconds when presented with a question. 
Fig. \ref{fig:Code_example} provides examples of SQL code generation for questions derived from a table. 
In Fig. \ref{fig:Code_example} (a), the prompt encompasses the table information and the instruction, while Fig. \ref{fig:Code_example} (b) and (c) present two questions along with their corresponding answers generated by LLMs. 
These examples demonstrate that SQL code can be effectively generated using high-level questions and prompts. 
The generated code can incorporate dynamic elements, such as converting "Hong Kong" to "HK"as demonstrated in Fig. \ref{fig:Code_example} (b), and understanding concepts like "Total" by utilizing "SUM" to aggregate multiple rows, along with the application of filters such as "where" and "like" to refine the result set. 
For a more detailed exploration of the code generation process, please refer to the subsequent section.

\begin{figure*}
\centering 
\includegraphics[width=5.5in]{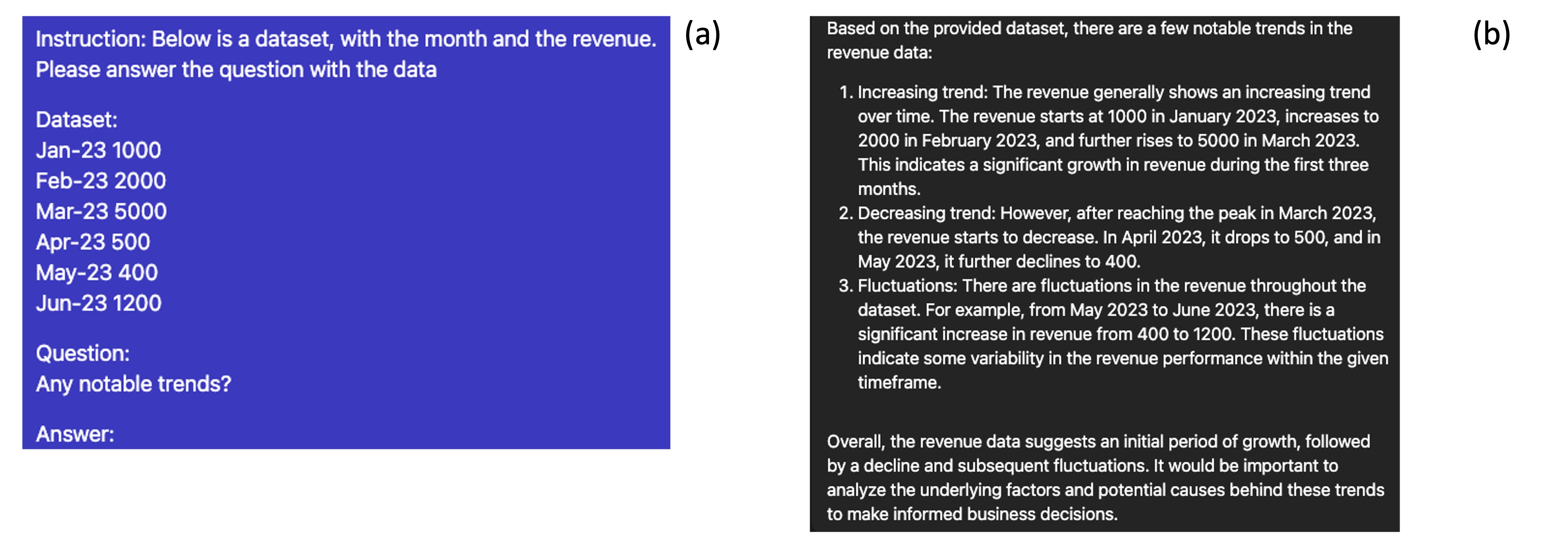} 
\caption{An example of data analytics using LLM: (a) The prompt with data, (b) The analytics.}
\label{fig:data_analytics_exp}
\end{figure*}

\subsection{Data Analysis}
Data analysis plays a crucial role in project planning and decision-making processes. While code generation focuses on extracting data, data analysis aims to comprehend and uncover valuable insights from the data. 
In the context of planning, data analysis can be utilized to validate whether implementing a recommendation engine would lead to increased sales. 
When it comes to decision making, data analysis enables the extraction of valuable insights and identification of trends from vast amounts of data, aiding in decisions such as targeting specific customer groups. 
These data-driven insights help businesses identify new market opportunities and remain competitive in a dynamic landscape. 
However, comprehending data requires expert insights and domain knowledge, as important patterns can be overlooked when dealing with large datasets. 
This is where LLMs excel; they are capable of extracting and analyzing meaningful information from datasets, such as alerting when sales drop. Fig. \ref{fig:data_analytics_exp} presents an example of a prompt, including the instruction, dataset, and question, along with the results provided by LLMs. 
The dataset used in this example is the monthly revenue, and the question posed is "Any notable trends?".
The results demonstrate that LLMs can uncover various dimensions of the data, highlighting periods of revenue increase, decrease, and a notable observation: revenue fluctuations. 
This showcases the potential of LLMs in assisting with data analytics. 
For a more detailed exploration of data analytics and a related experiment, please refer to the next section.

\begin{figure*}
\centering 
\includegraphics[width=5.5in]{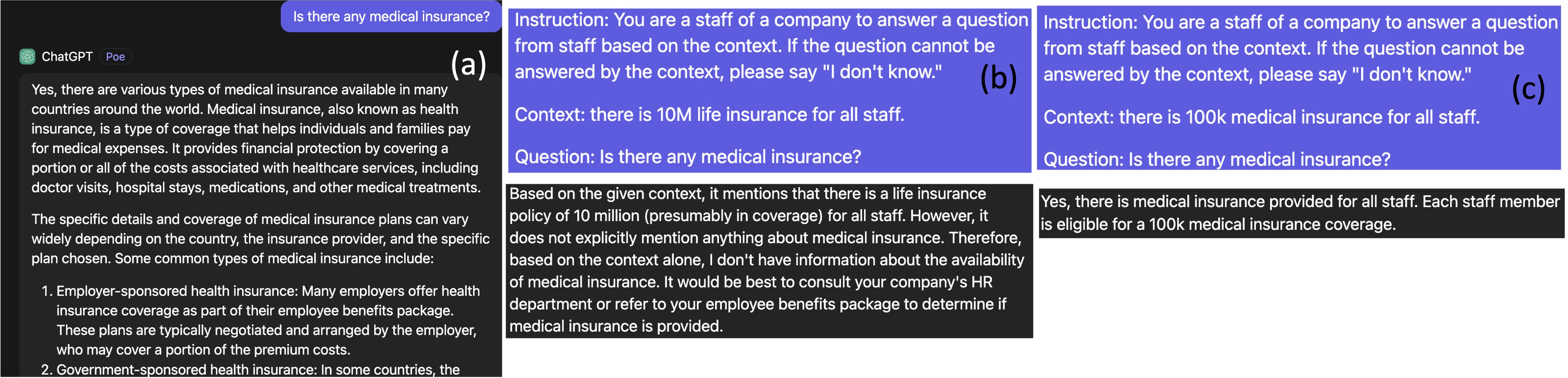} 
\caption{An example of a question using context: (a) without context, (b) with a wrong context, (c) with the right context.}
\label{fig:Context_QA_exp}
\end{figure*}

\subsection{Chatbot with Contexts}
\noindent
LLMs form a solid foundation for the development of chatbots or virtual assistants that can effectively interact with users and provide informative responses \cite{anil2023palm}. 
These models possess impressive text comprehension capabilities, enabling them to generate relevant answers based on user input. 
However, instead of relying solely on generative capabilities, LLMs can be instructed to answer questions based on specific contexts provided by users.
The importance of context in generating accurate responses can be illustrated through an example depicted in Fig. \ref{fig:Context_QA_exp}. 
In Fig. \ref{fig:Context_QA_exp} (a), a question about medical insurance is asked without any context, resulting in a general response about different types of insurance. 
However, this generic response may not cater to the specific requirements of a particular company. 
Fig. \ref{fig:Context_QA_exp} (b) demonstrates a scenario where the provided context lacks sufficient information to answer the question, leading the LLM to indicate the inadequacy of information. 
Conversely, in Fig. \ref{fig:Context_QA_exp} (c), the LLM is able to provide a correct and relevant answer due to the presence of suitable context.
This example highlights the ability of LLMs to offer customized responses beyond their training set when provided with appropriate context. 
Whether it's a given document, such as a contract, or a personal assistant catering to customer inquiries, LLMs can leverage the provided context to provide accurate and relevant answers. 
Consequently, incorporating the right context within prompts becomes crucial for LLMs to function effectively.
\\
\indent
The upcoming sections will explore the limitations of LLMs, including bias, a lack of context understanding, and sensitivity to prompts. The performance and effectiveness of various LLMs will be evaluated in different use cases and assess their limitations, such as reference generation, code generation, and Q\&A from context, through experiments. 
For the experiments, multiple accessible LLM models are tested through Poe\footnote{https://poe.com/}, a chatbot wrapper that enables users to access various web-based models using prompts. 
To automate the process of capturing responses, a bot is developed to call the required models. 
Further details about the experiment can be found in the subsequent sections.

\section{Bias}
\label{sec:bias}
\noindent
This section explores the impact of bias on the application of LLMs in business settings. 
LLMs are susceptible to various forms of bias, including gender bias \cite{borji2023categorical}, popularity bias, recency bias \cite{zhao2021calibrate}, and linguistic bias, which can lead to underperformance compared to traditional recommender systems \cite{zhang2021language}. 
Moreover, biases can manifest in the unequal representation of digits, with certain digits appearing more frequently than others \cite{azaria2022chatgpt}. 
These biases primarily arise from biased training data and the fine-tuning process. 
Limited diversity in training data can contribute to the acquisition and perpetuation of biases, while fine-tuning can further amplify existing biases \cite{renduchintala2021gender}. 
When the data primarily represents a specific demographic, LLM-generated content may reinforce biases against marginalized communities. 
The presence of bias is a significant concern as it not only provides customers with incorrect information but also negatively impacts a company's reputation. 
These challenges underscore the importance of ongoing research and improvement to address the limitations associated with bias. 
To assess the impact of bias on business planning, an experiment has been conducted to investigate the usefulness of references suggested by LLMs. 
This experiment simulates a user initiating a project, utilizing LLMs to gain an understanding of a new field by studying state-of-the-art papers and grasping the fundamental concepts within that field.

\subsection{Settings}
\noindent
The goal of the experiment is to generate reference papers based on a topic.
In the experiment, a reference dataset comprising five survey papers is collected, with each survey's references serving as the ground truth.
These surveys were selected to represent a diverse range of topics and domains and were published in 2022. 
Some topics are popular, such as transformers.
On the other hand, applying federated learning in the medical area, is less popular, even that this paper has more citations.
The five surveys are as follows (citation count as of Nov 2023, from Google Scholar):
\begin{itemize}
\item Wen, Jie, et al. "A survey on incomplete multiview clustering." IEEE Transactions on Systems, Man, and Cybernetics: Systems 53.2 (2022): 1136-1149. Cited 55 times.
\item Gupta, Jaya, et al. "Deep learning (CNN) and transfer learning: a review." Journal of Physics: Conference Series. Vol. 2273. No. 1. IOP Publishing, 2022. Cited 25 times.
\item Liu, Yang, et al. "A survey of visual transformers." IEEE Transactions on Neural Networks and Learning Systems (2023). Cited 123 times.
\item Goudarzi, Arman, et al. "A Survey on IoT-Enabled Smart Grids: Emerging, Applications, Challenges, and Outlook." Energies 15.19 (2022): 6984. Cited 59 times.
\item Nguyen, Dinh C., et al. "Federated learning for smart healthcare: A survey." ACM Computing Surveys (CSUR) 55.3 (2022): 1-37. Cited 219 times.
\end{itemize}
In the experiment, the titles of the papers are used as prompts for the four LLMs, which suggest 50 paper titles each. 
The experiment measures the number of LLM-suggested papers that exist in the reference list of the surveys, serving as an indicator of whether an LLM can effectively suggest papers for project planning. 
Note that the matching of suggested papers with the reference list considers only the titles, as they are sufficient for identifying the papers, while LLMs tend to generate errors in the author list.

\begin{figure*}
\centering 
\includegraphics[width=5.5in]{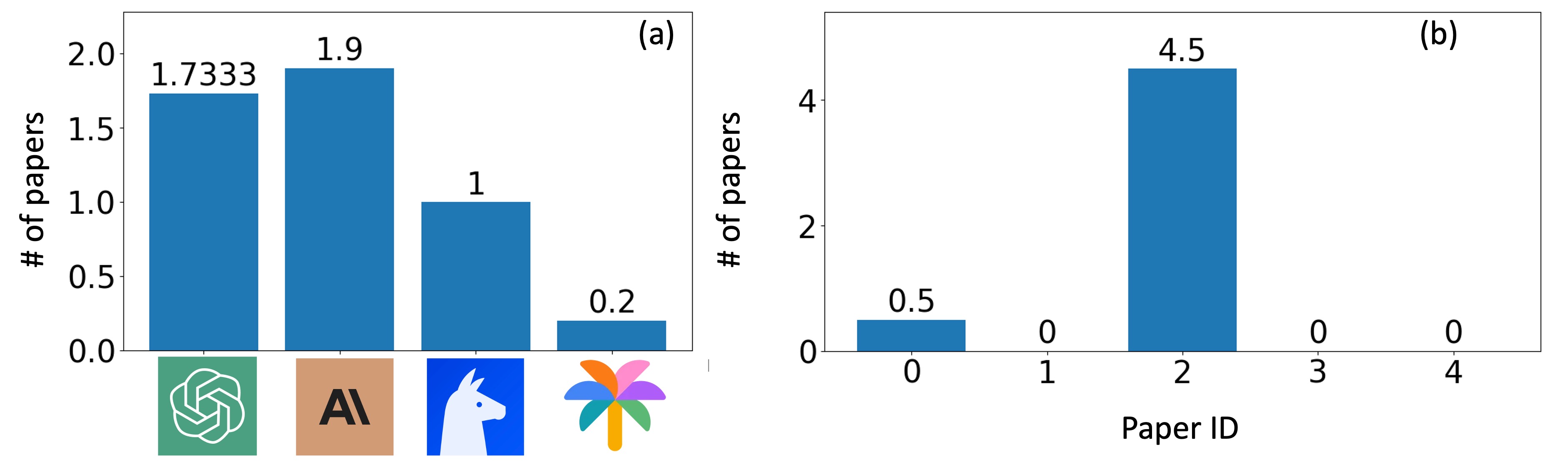} 
\caption{Result of reference generation, in no. of papers: (a) by LLMs, (b) by the five papers.}
\label{fig:e1_result}
\end{figure*}

\begin{figure*}
\centering 
\includegraphics[width=5.5in]{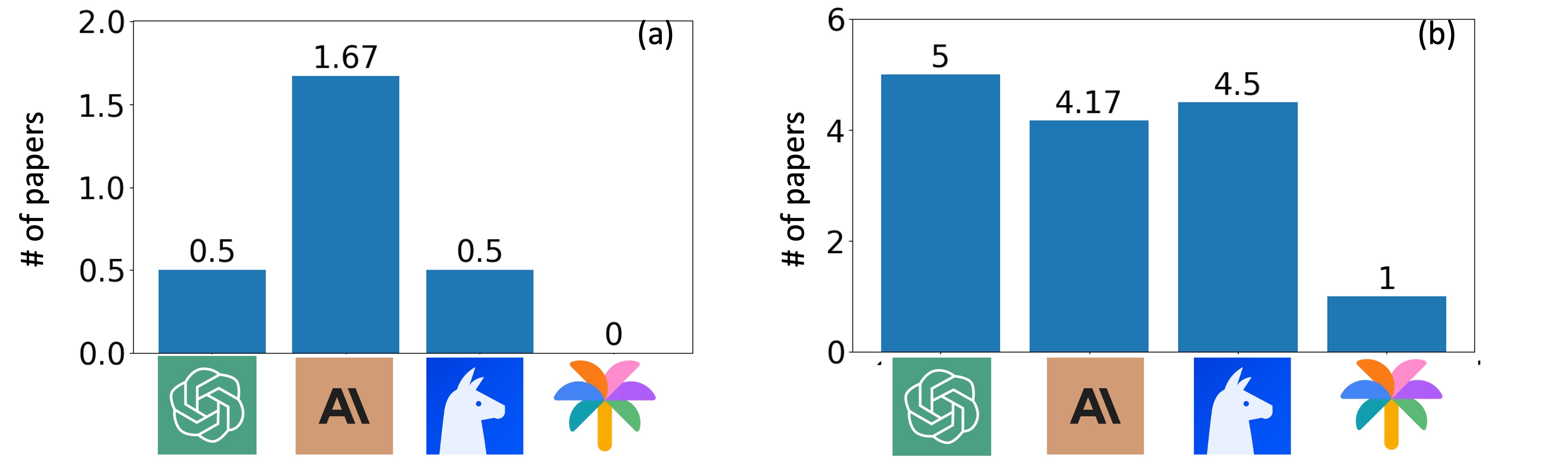} 
\caption{Result of: (a) paper 0, (b) paper 2.}
\label{fig:e1_result_papers}
\end{figure*}

\subsection{Results}
\noindent
Fig. \ref{fig:e1_result} presents the results of the experiments. 
Fig. \ref{fig:e1_result} (a) shows the mean number of suggested papers from the LLMs that are present in the surveys. 
It is observed that Claude-instant performs the best, with an average of 1.9 papers out of the 50 generated papers actually being in the surveys.
ChatGPT follows closely, with 1.73 papers. 
However, even for the best-performing LLMs, only a few papers match the surveys. 
This could be attributed to LLM limitations in handling factual information. 
Fig. \ref{fig:e1_result} (b) displays the mean number of generated papers that exist in the reference list of each survey. 
It is observed that the third paper, "A Survey of Visual Transformers," has the highest number of matches. 
This is likely because there are numerous samples related to Visual Transformers, making it easier for LLMs to suggest papers in that area. 
Conversely, the other papers, being less popular, have fewer generated papers that match the surveys. 
The results highlight that LLMs may unknowingly learn and perpetuate biases in the training dataBased on the conducted experiment, it was found that LLMs have limitations in suggesting relevant papers for project planning. 
The best-performing LLM in the experiment, Claude-instant, only had an average of 1.9 suggested papers out of 50 that matched the reference list of the surveys. 
One of the reasons is that LLMs can unknowingly learn and perpetuate biases in the training data, leading to biased or unfair generations.
Fig. \ref{fig:e1_result_papers} (a) and Fig. \ref{fig:e1_result_papers}(b) show the results for paper 0 and paper 2 on different LLMs.
They are the 2 papers with matched generated papers.
It is observed that in Fig. \ref{fig:e1_result_papers}(b), the most popular paper, the three LLMs: ChatGPT; Claude; and Llama, they show very similar performance, while in Fig. \ref{fig:e1_result_papers}(a), Claude outperformes others.

\subsection{Discussion}
\noindent
The experiment revealed that the popularity of a topic can impact the number of matching papers suggested by LLMs. 
The survey paper on "Visual Transformers" had the highest number of matches, in which there are over 300k results on Google Scholar for searching "Visual Transformers".
It is a popular topic with a larger pool of related papers available.
In contrast, less popular topics had fewer matching papers suggested by LLMs.
Searching "Federated learning for healthcare" on Google Scholar found less than 90K results.
Therefore, a crucial research direction is focused on mitigating bias within training data\cite{bordia2019identifying}. 
One approach to measure bias in datasets and resulting models is by utilizing common demographic identity terms\cite{dixon2018measuring}. 
To ensure the accuracy of inputs in LLMs, it is imperative to establish a thorough review and evaluation process\cite{roselli2019managing}. 
Furthermore, enhancing the transparency and accountability of LLMs can be achieved by providing comprehensive documentation of model architectures, training data, and fine-tuning processes\cite{perez2021true}. 
This documentation enables the evaluation of potential sources of bias. 
Another direction for reducing bias involves the utilization of appropriate prompts\cite{mallen2023not}. 
By providing additional context on less popular content, bias can be minimized, and incorporating translations can offer diverse perspectives\cite{ghazvininejad2023dictionary}. 
In the context of paper generation, using search results from Google Scholar as input for LLMs can contribute to bias reduction.

\section{Lack of Context Understanding}
\label{sec:no_context}
\begin{figure*}
\centering 
\includegraphics[width=5.5in]{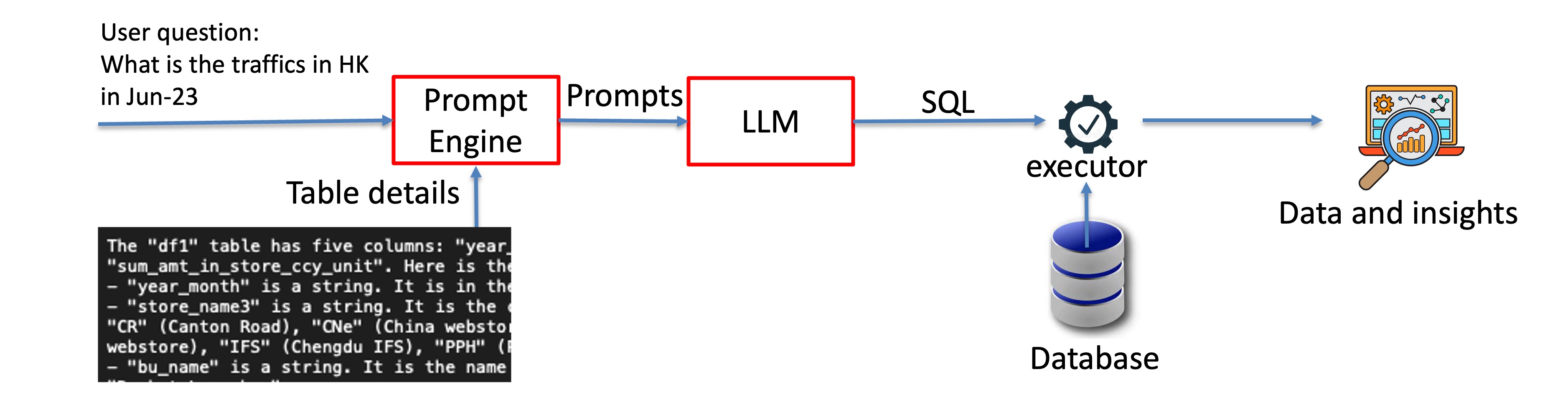} 
\caption{The flow for generating SQL code using LLM.}
\label{fig:Code_gen}
\end{figure*}

\noindent
This section examines the impact of the lack of context understanding in LLMs on business applications. 
While LLMs are capable of generating fluent text, they often lack factual accuracy \cite{Stokel2023what}. 
Particularly in complex problem-solving scenarios, such as those involving mathematics and coding, LLMs struggle due to the need for domain-specific knowledge and context comprehension. 
This limitation restricts their effectiveness in domains that rely on robust numerical reasoning, as evidenced by their difficulties in mathematical tasks \cite{azaria2022chatgpt}. 
One of the reasons behind this challenge is that LLMs are trained to generate subsequent words based on prompts and previously generated text, rather than comprehending the full context. 
Consequently, it is worth investigating how this limitation affects complex tasks like coding.
\\
\indent
Structured Query Language (SQL) is a widely used programming language for data extraction from databases, making it valuable for various business applications. 
However, effectively utilizing SQL often requires the expertise of a business analyst. 
This process can be time-consuming, especially when prompt-specific insights require further exploration. 
For instance, when a business user wants to delve deeper into identified patterns, they may have to wait for the next meeting, thereby slowing down the decision-making process. 
Leveraging LLMs to generate SQL queries from natural language can expedite the exploration of insights, thereby accelerating the decision-making process.
\\
\indent
Fig. \ref{fig:Code_gen} illustrates the flow of the prompt for generating SQL queries from questions. 
By providing a prompt with the necessary table details, an SQL code can be generated and executed to retrieve data from the tables. 
This data can then be further evaluated and used to address user inquiries. 
For example, a question like "What is the conversion rate in Hong Kong?" would require extracting relevant data using SQL and calculating the rate accordingly. 
To experiment with generating code for extracting required data based on business questions, an experiment is conducted.

\begin{table}[]
\caption{Table details.}
\begin{tabular}{llll}
Name         & Explanation                   & Tables & Examples        \\
Date         & date, in format of YYYY-MM    & d1, d2 & 2022-03         \\
Store        & Store Name                    & d1, d2 & IFC             \\
Region       & Region Name                   & d1, d2 & HK              \\
Business     & The name of the business unit & d1     & WW (woman wear) \\
Amount       & The sum of transaction amount & d1     & 1400            \\
Traffic      & The number of traffic         & d2     & 10,000          \\
Transactions & The number of transactions    & d2     & 1,000          
\end{tabular}
\label{tab:details}
\end{table}

\subsection{Experimental Settings}
\noindent
This experiment aims to assess the accuracy of various LLMs in generating SQL code for given questions and obtaining relevant results. 
The experiment utilizes a dataset consisting of two tables: d1 (sales data) and d2 (traffic data) of stores.
The first table, d1, stores the cumulative sales of different business units in a store, while the second table, d2, stores the number of traffic and transactions in a store. 
Details of these tables are provided in Tab. \ref{tab:details}. 
It is worth noting that certain columns exist exclusively in one of the tables, such as the business unit and the amount.
Some questions require data from both d1 and d2, as they involve calculations using columns from both tables. 
For instance, determining the average order value of a region or store would necessitate data from both d1 and d2. 
This can be computed by dividing the column "Amount" in d1 by the column "Transactions" in d2. Consequently, joining the two tables is necessary for such calculations.
\\
\indent
A set of 33 analytics questions, along with their expected results, is prepared for the experiment. 
These questions fall into three categories. 
The first category includes questions that can be answered using only d1, such as "What is the total sales for the month of July 2022?". 
The second category comprises questions that solely require d2. 
Lastly, the third category, d1\&d2, involves questions that require both tables. 
An example of a question from this category is "Which region had the highest AOV in February 2022?". 
Here, AOV refers to Average Order Value and can be calculated by dividing the "Amount" column in d1 by the "Transactions" column in d2, under the condition that the "Date" and "Region" columns are the same. 
This category is more complex compared to the first two.
\\
\indent
The performance of the experiment is evaluated based on accuracy, match rate, and run rate of the generated SQL.
Accuracy refers to whether the query result matches the ground truth. This criterion is particularly challenging as it requires the SQL to run correctly and extract data from the appropriate tables. 
While accuracy is crucial, indicators such as whether the SQL retrieves data from the correct table and whether the SQL can run successfully are also considered. 
The match rate is measured as the percentage of SQL queries that retrieve the required data from the correct tables, regardless of the correctness of the answers. 
Similarly, the run rate is measured as the percentage of SQL queries that can run, regardless of the correctness of the answers or tables involved.
The results of the experiment are presented in the subsequent section.

\begin{figure*}
\centering 
\includegraphics[width=5.5in]{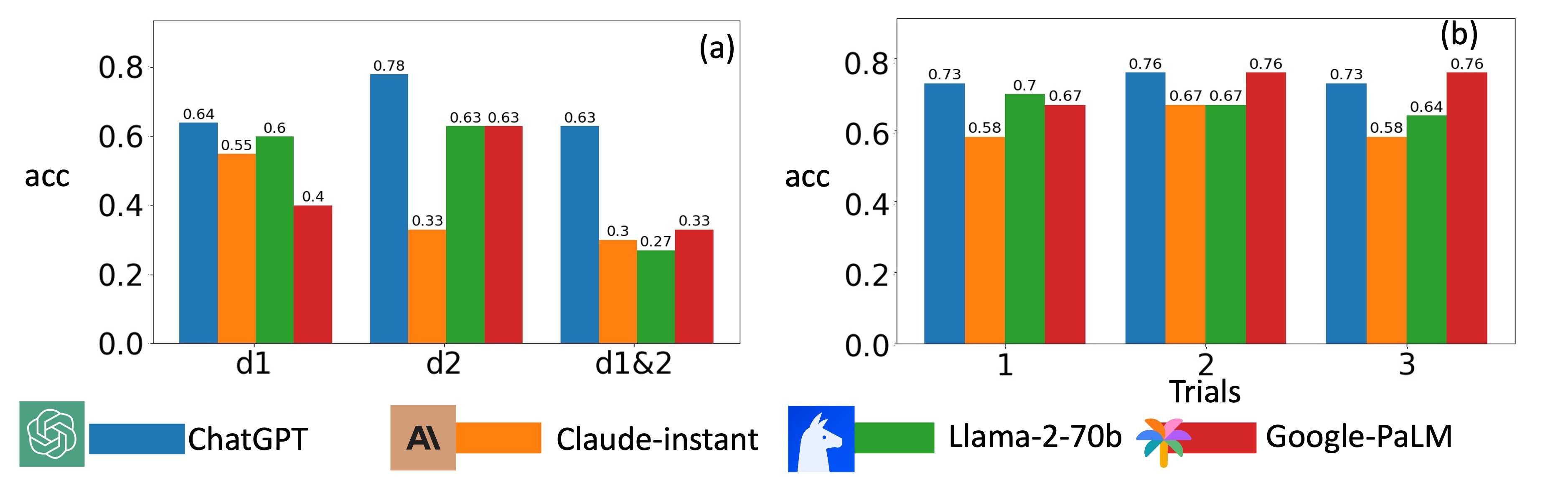} 
\caption{Result of code generation, in accuracy: (a) by tables required, (b) by the trials.}
\label{fig:e2_result}
\end{figure*}


\begin{figure*}
\centering 
\includegraphics[width=5.5in]{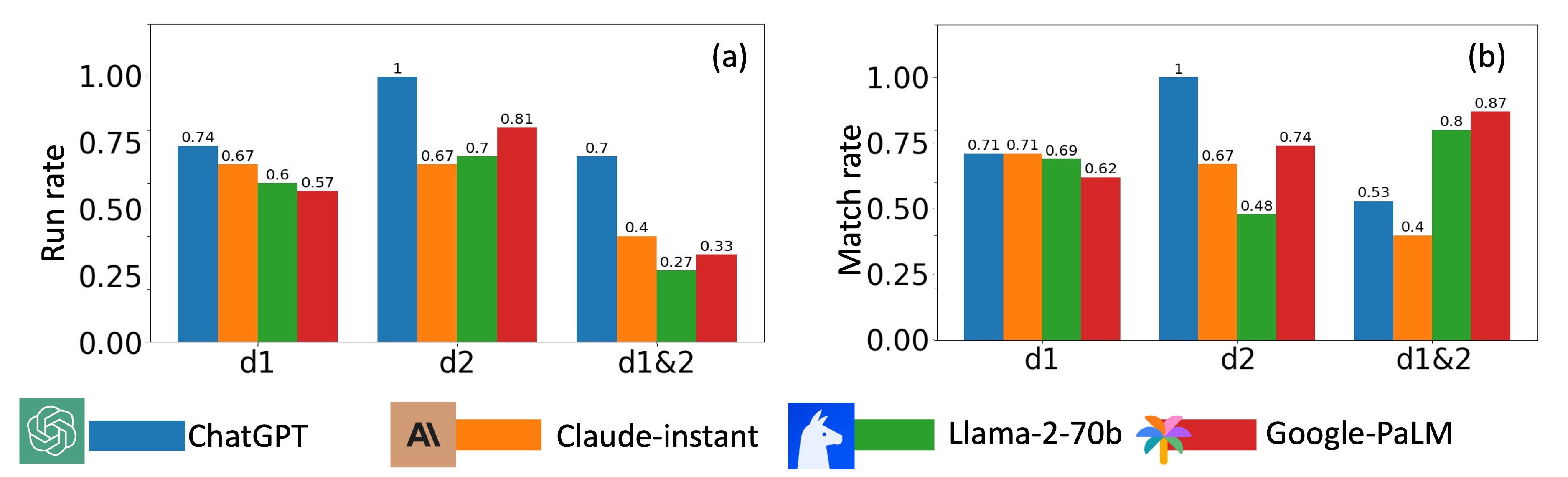} 
\caption{Result of code generation by the tables required, in: (a) run rate, (b) match rate.}
\label{fig:e2_run_table}
\end{figure*}

\subsection{Results}
\noindent
Fig. \ref{fig:e2_result} presents the accuracy of the generated SQL from questions. 
In Fig. \ref{fig:e2_result} (a), the SQL is grouped based on the tables required to generate it, namely, d1, d2, or d1\&d2. 
It is observed that across all LLMs, SQL queries that only involve a single table perform better. 
Among the four LLMs, ChatGPT demonstrates the highest performance. 
It is important to note that these results represent the mean of three trials. 
Additionally, it is interesting to explore the impact of multiple independent queries for the same question on performance. 
Fig. \ref{fig:e2_result} (b) illustrates the results for different trials, revealing that multiple queries affect the consistency of accuracy as LLMs generate different SQL queries for the same questions. 
Fig. \ref{fig:e2_run_table} (a) and (b) display the run rate and match rate, respectively, for different question types. 
It is observed that ChatGPT predominantly produces runnable SQL queries for single table cases, and the matching rate is consistently high across LLMs. 
Questions that only require d1 generally yield better results in ChatGPT, which d1\&d2 in Google-PaLM.
One of the results is that Google-PaLM produces many more times that the generated SQL contains both d1 and d2, even that it is required data from 1 table only.
To summarize, ChatGPT demonstrates the best performance, but LLMs struggle with joins that necessitate more complex SQL queries.

\subsection{Discussion}
\noindent
The results depicted in Fig. \ref{fig:e2_result} and Fig. \ref{fig:e2_run_table} indicate that LLMs excel at comprehending questions and generating SQL queries that do not involve utilizing both tables. 
However, handling joins, which are more intricate, is better suited for human intervention. 
This is because LLMs lack an understanding of the context. 
Therefore, a promising research direction is the development of context-aware LLMs that leverage contextual information \cite{anand2023context}. 
The objective is to devise methods that can comprehend and utilize context cues, such as previous conversation history or external knowledge, to generate more accurate and contextually appropriate responses. 
Another research avenue involves evaluating the accuracy of code generation \cite{kashefi2023chatgpt}. 
In code generation tasks, it is challenging to detect errors in the SQL code since users can only assess the final output if the SQL query successfully executes. 
This limitation hampers the reliability, which is crucial for LLMs to be truly valuable in a business setting. 
Therefore, an additional direction of research involves developing techniques to detect errors in the generated SQL queries and their corresponding answers.


\section{Sensitivity to Prompts}
\label{sec:sensitivity}
\begin{figure*}
\centering 
\includegraphics[width=5.5in]{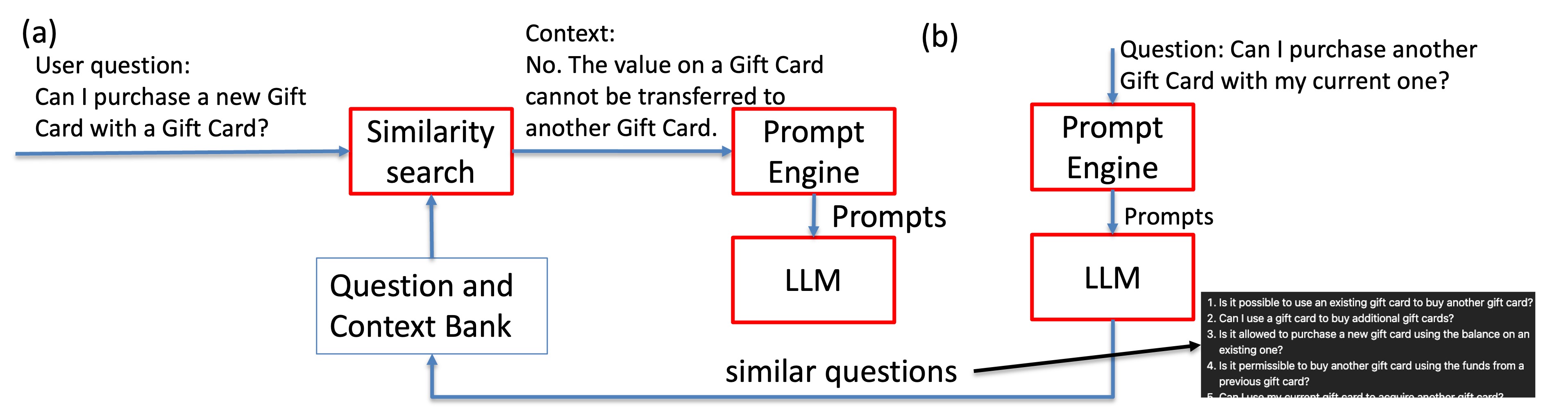} 
\caption{The flow for questions and answers with context: (a) answering a question, (b) generating questions.}
\label{fig:Context_QA}
\end{figure*}

LLMs have demonstrated remarkable capabilities, but they are also known for their sensitivity to the prompts used during inference \cite{xie2023translating, wang2022rationale}. 
This sensitivity poses a challenge when applying LLMs to business-related tasks. 
Notably, the choice of prompts can significantly impact the accuracy of generated results \cite{jang2023can}. 
Even a slight modification in the prompt can lead to substantial variations in the model's outputs \cite{arora2022ask}. 
Therefore, providing the appropriate context is crucial for many applications, as depicted in Fig. \ref{fig:Context_QA_exp}. 
However, extracting the right context for a given question is not always straightforward.
\\
\indent
One particular use case where context plays a vital role is answering customer questions on an e-commerce site, which often requires substantial human resources. 
Typically, companies prepare a set of question-answer pairs to address common queries. 
For instance, a pair may consist of the question, "Can I add additional items to my order?" and its corresponding answer, "Once your order has been submitted, you cannot add or remove the items in it.". 
When faced with new questions, one approach is to find the most similar question from the existing pairs and use the associated answer as the context. 
For instance, the question "How to remove an item from an order" can be answered using the above context, indicating that customers cannot remove items. 
LLMs can infer the answer from the given context. However, variations in wording and contextual meaning across questions can pose challenges.
\\
\indent
To address this issue, data augmentation is a useful solution for generating additional questions for each context. 
Fig. \ref{fig:Context_QA} illustrates a question and answer (Q\&A) system that employs data augmentation with LLMs. 
In Fig. \ref{fig:Context_QA} (a), the system follows a flow for question and answer retrieval with contexts. Starting with a user question, the system performs a similarity search to find similar questions from a question and context bank. 
The question and context bank consists of pairs of questions and contexts, where a single context can be associated with multiple questions. 
Once the similar questions are identified, the corresponding context is extracted and used to answer the question using LLMs with prompts. 
However, obtaining a comprehensive question bank is often challenging. For instance, in a company's Q\&A section, there may be only one question matched to one context. 
To address it, Fig. \ref{fig:Context_QA} (b) presents a flow for extending the question and context bank using LLMs. 
The prompts guide the LLMs to generate similar yet distinct questions based on existing examples and the context. 
By augmenting the data, the accuracy of the similarity search is improved. 
In the following section, an experiment is conducted to investigate whether data augmentation using existing questions can enhance the performance of LLMs in question answering.

\subsection{Settings}
\noindent
The dataset used in this study comprises pairs of questions and answers obtained from an e-commerce website\footnote{available: https://www.lanecrawford.com.hk/info/help/help-and-info/}.
The original dataset consists of 20 questions, each accompanied by its corresponding answer, which serves as the context for answering the questions.
To create a larger and more diverse dataset, data augmentation techniques using LLMs were applied to the questions.
For each prompt, 20 augmented questions were generated, and this process was repeated 6 times.
It is important to note that each prompt is independent, and the questions from previous prompts are not included in the input.
In total, each LLM is expected to produce 2400 questions.
However, as some questions could be repeated, the final number of questions are less than 2400 questions.
The augmentation was performed using all 4 LLMs, and the augmented data was tested separately.
In the experiment, the questions from the first to fifth trials were used as the training set, that is, the question set to be compared with a query question.
The questions from the sixth trial of all 4 LLMs were used as the testing set, that is, the question set to be used as the query questions.
This ensures that the questions are not biased towards a particular LLM.
Once the questions were prepared, they were encoded using Sentence-BERT \cite{reimers2019sentence}.
The Similarity Search algorithm was applied, using $k$-nearest neighbors ($K$-NN) with $k=5$, to identify the context with the highest similarity to each question.
The performance of the LLMs was evaluated based on accuracy (acc), which measures the percentage of testing questions that were correctly matched with the appropriate context.
\\
\indent
In addition to comparing the performance of the LLMs, various approaches for sample augmentation were also compared to identify the most effective method for improving the performance of the LLMs.
The first approach, referred to as LC, involved using scraped data as the 1-NN in the similarity search.
The other two approaches were implemented using nlpaug\footnote{available: https://nlpaug.readthedocs.io/}.
The second approach, referred to as Ctx, utilized the context within the questions by using the "bert-base-uncased" model, along with the "insert" and "substitute" operations.
The third approach, referred to as Rand, involved randomly swapping words in the questions.
An equal number of training samples were generated for both Ctx and Rand approaches.

\begin{figure*}
\centering 
\includegraphics[width=5.5in]{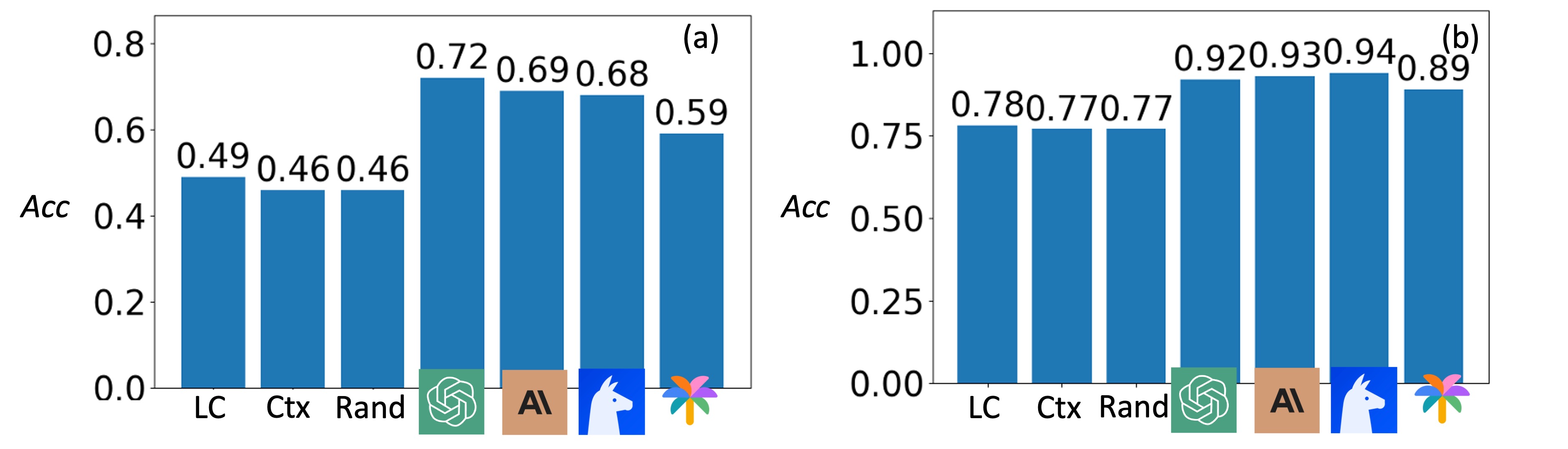} 
\caption{Result of context extraction, in accuracy and by algorithms and LLMs, (a) top 1, (b) top 3.}
\label{fig:e3_knn}
\end{figure*}

\subsection{Results}
The results indicate that LLMs effectively enhance data for extracting the relevant context. 
Fig. \ref{fig:e3_knn} (a) and (b) illustrate the outcomes of different augmentation methods in terms of top 1 and top 3 accuracy, respectively. 
Considering that LLMs can accommodate multiple contexts in a single prompt and generate answers accordingly, the top 3 contexts are also considered acceptable.
Notably, the samples generated by LLMs yield superior results, outperforming the use of Q\&A data alone. 
Conversely, the other two augmentation methods exhibit limited effectiveness, as they do not significantly influence the accuracy. 
These findings highlight the capability of ChatGPT to yield the most improvement and generate valuable samples.

\begin{figure*}
\centering 
\includegraphics[width=5.5in]{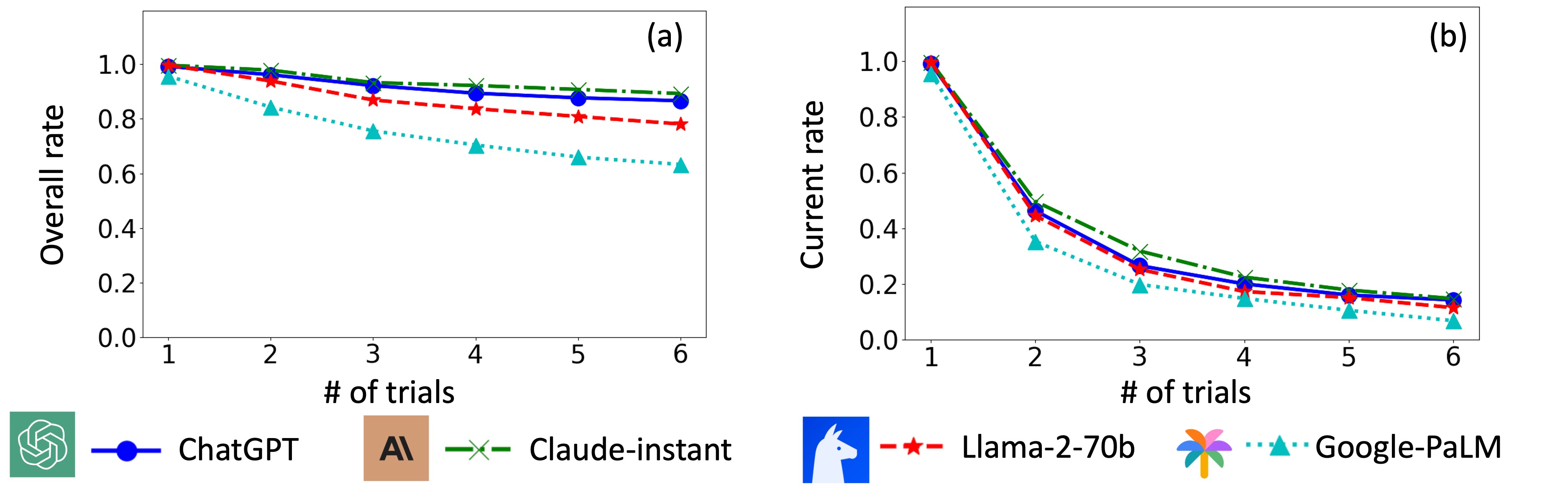} 
\caption{Uniqueness rate with the number of trails, (a) overall, (b) current.}
\label{fig:e3_unique}
\end{figure*}

Additionally, aside from accuracy, an important comparison is whether conducting more trials can lead to the generation of new questions. 
Due to the inherent randomness in generation, LLMs may produce different answers even when the same questions and prompts are employed. 
Fig. \ref{fig:e3_unique} presents the overall and unique question percentages in relation to the number of trials. 
The overall rate represents the percentage of questions that are unique when considering all questions from the first trial. 
The current rate signifies the percentage of questions that are unique and do not exist in the previous trial. 
The results demonstrate that ChatGPT and Clande achieve the best performance in this regard.

\subsection{Discussion}
As demonstrated in Fig. \ref{fig:e3_knn}, LLMs can be effectively utilized for question augmentation in Q\&A systems. 
This observation leads to several potential research directions. 
Firstly, it is worth investigating the impact of prompts on generated responses to mitigate undesired behaviors resulting from prompt manipulation. 
Techniques such as providing samples \cite{arora2022ask} and using templates \cite{mishra2021reframing} can be explored. 
Different prompt formats, including prompt length, style, question-answering formats, and context inclusion/exclusion, can be experimented with to maximize task performance.
Secondly, the generation of prompts using models or existing data can be explored to improve performance on specific tasks. 
Reinforcement learning can be leveraged as an efficient approach for discrete prompt optimization \cite{deng2022rlprompt}, where the reward of a prompt is predicted to propose the next one.
Lastly, the transfer learning of prompts can be investigated to transfer knowledge learned from one task to another through prompts. 
Methods that leverage task similarity, data, and prompts from related tasks \cite{cheung2023learning} can be explored. 
Research on encoding task data and prompts can prove beneficial in enhancing prompts.

\section{Conclusion}
\label{sec:conclusion}
\noindent
In conclusion, this paper presents a pioneering quantitative study that evaluates the practicality of LLMs beyond their traditional applications in language understanding and generation tasks, specifically focusing on their usefulness in common business processes.
By conducting experiments using enterprise data, the authors comprehensively assess the performance of LLMs in planning, implementation, and decision-making tasks. The study provides initial insights into the areas where LLMs can enhance human work and identifies their current limitations.
The results demonstrate that LLMs can be valuable in coding and Q\&A tasks, but they exhibit limitations in terms of bias, contextual understanding, and sensitivity to prompts.
These findings hold significant implications for organizations seeking to harness the power of generative AI, while also shedding light on ongoing research opportunities to expand the adoption of LLMs in professional settings.
Overall, this research advances our understanding of the practical applications of LLMs and offers valuable insights for effectively integrating these models into core business operations.
\bibliographystyle{ACM-Reference-Format}
\bibliography{llm}
\end{document}